# Hybrid Attention-Based Transformer Block Model for Distant Supervision Relation Extraction

Yan Xiao [a], Yaochu Jin [a,b,*], Ran Cheng [c], Kuangrong Hao [a],

[a] *Engineering Research Center of Digitized Textile & Apparel Technology, Ministry of Education, College of Information Science and Technology, Donghua University, Shanghai 201620, P. R. China*
[b] *Department of Computer Science, University of Surrey,Guildford, Surrey GU27XH, United Kingdom*
[c] *The Shenzhen Key Laboratory of Computational Intelligence, University Key Laboratory of Evolving Intelligent Systems of Guangdong Province, Department of Computer Science and Engineering, Southern University of Science and Technology, Shenzhen 518055, China.*

A B S T R A C T

With an exponential explosive growth of various digital text information, it is challenging to efficiently obtain specific knowledge from massive unstructured text information. As one basic task for natural language processing (NLP), relation extraction aims to extract the semantic relation between entity pairs based on the given text. To avoid manual labeling of datasets, distant supervision relation extraction (DSRE) has been widely used, aiming to utilize knowledge base to automatically annotate datasets. Unfortunately, this method heavily suffers from wrong labelling due to the underlying strong assumptions. To address this issue, we propose a new framework using hybrid attention-based Transformer block with multi-instance learning to perform the DSRE task. More specifically, the Transformer block is firstly used as the sentence encoder to capture syntactic information of sentences, which mainly utilizes multi-head self-attention to extract features from word level. Then, a more concise sentence-level attention mechanism is adopted to constitute the bag representation, aiming to incorporate valid information of each sentence to effectively represent the bag. Experimental results on the public dataset New York Times (NYT) demonstrate that the proposed approach can outperform the state-of-the-art algorithms on the evaluation dataset, which verifies the effectiveness of our model for the DSRE task.

## 1. Introduction

In the age of information, the amount of data that people need to process has surged constantly due to the rapid development of internet technology. How to extract effective information from these open-field texts quickly and efficiently has become an important issue nowadays. As the core task of text mining and information extraction [7], the results of relation extraction (RE) are mainly applied to many NLP applications, which is designed to generate relational data from plain text. To be specific, the relation of an entity pair can be formalized as a relational triple $\langle e_1, r, e_2 \rangle$, where $e_1$ and $e_2$ are entities and $r$ implies the relation between two entities in the sentence. The task of RE aims to extract the relational triple $\langle e_1, r, e_2 \rangle$ from the natural language text, consequently achieving the information extraction.

There are various approaches proposed for RE, among which the supervised paradigm has been demonstrated to be effective and exhibited relatively high performance. However, the supervised method always requires to manually collect large amount of labeled data, which is time-consuming and labor-intensive. Meanwhile, there are various abundant knowledge bases (KBs) such as Freebase [5], DBpedia [1] and Wikidata [40] which have been built and widely used in many NLP tasks as data resources. Thus, to solve the problem of manually annotating datasets, Mintz et al. [25] proposed the distant supervision relation extraction (DSRE) which can automatically generate training data via aligning triples in KBs with sentences in corpus. The method assumes that if two entities have a relation in KBs, then all sentences that contain these two entities will describe the same relation. As illustrated in the Fig. 1, according to the triple < *Kai-fu Lee, /business/person/company, Microsoft* >, all the sentences mentioning "*Kai-fu Lee*" and "*Microsoft*" will be labeled as the training instances of relation "*/business/person/company*", despite that not all sentences express the same relation.

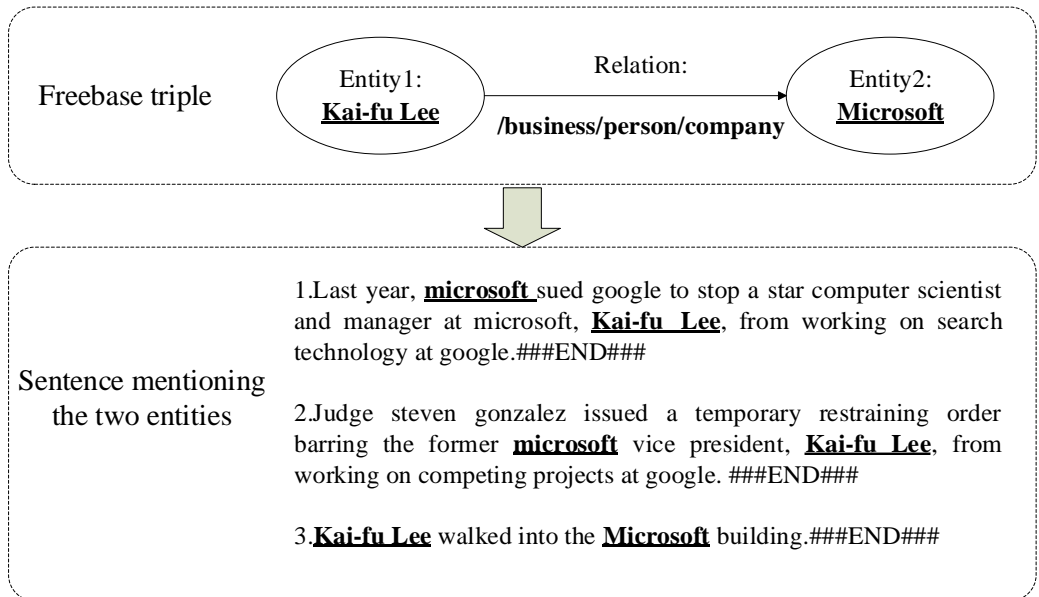

**Fig. 1.** Training sentences generated through distant supervision by aligning entity tuples in a text corpus with relation facts from a knowledge base, where the third sentence is mislabeled.

Therefore, although the distant supervision strategy is an effective way to automatically label the data, it also brings a new issue: it may be subject to noisy labels due to the fact that not all sentences containing two entities necessarily express the same relation in KBs. The third sentence in the Fig. 1 is a good example of this issue. More recently, researchers have adopted multi-instance learning to alleviate this issue, one of which relaxes the assumption to the at-least-one principle [34]. The principle argues that if two entities participate in a relation, there will be at least one sentence mentioning these entities implies the relation. Accordingly, the dataset of DSRE is utilized for training models on bag-level, which implies that all sentences containing the same entity $e_1$ and $e_2$ are packaged into one bag, and $r$ is the relation label of the bag, denoted as $\langle e_1, r, e_2 \rangle \{S_1, S_2, \cdots, S_n\}$. The sentences in each bag are also called instances. Then the major task of DSRE is to predict the corresponding relation label $r$ for each bag that consists of sentences with the same entity pairs. Besides, this task can be applied to various real-world scenarios, since it is capable of determining whether two entities have a corresponding relationship.

Furthermore, previous traditional distant supervision models typically applied supervised models to handcrafted features, where most features are explicitly derived from NLP tools such as POS tagging. Hence, the errors generated by such tools will propagate and accumulate, resulting in performance degradation. With the rapid advances of deep learning in recent years, researchers have attempted to utilize deep neural networks for relation extraction, which has achieved significant performance improvement. Compared with the conventional models, deep neural networks allow to efficiently extract semantic features of sentences without any tools or handcrafted features. The basic frameworks of most existing DSRE models are mainly based on Piecewise Convolutional Neural Network (PCNN) [32,48,49] or Long Short Term Memory networks (LSTM) [44,51], where CNN is able to extract local features and LSTM is able to extract global features of a sequence. However, it has also been widely recognized that CNN is restricted to solving the problem of long and short distance dependence in sequence, while LSTM has poor concurrence. In addition, both of them are weak in capturing the global dependency of a long sequence, which indicates that learning a good sentence encoder remains challenging. Moreover, most current studies in the field of DSRE have adopted the attention mechanism [22] to fuse information for each entity pair in the process of dealing with wrong labels, which is computationally intensive during test due to the embedding of relation. It may cause misjudgments if there are two labels expressing the similar meanings, which calls for a more effective bag presentation.

To get a more extensive vector expression for each sentence and better address the wrong labelling problem, this work proposes a hybrid attention-based Transformer block model to improve the performance of DSRE, called Trans-SA for short. As illustrated in Fig. 2, the overall model consists of two parts: Transformer block -based sentence encoder and sentence-level attention-based bag representation. The first part plays the same role as PCNN or LSTM as it is used to encode each sentence of the input. It is in principal based on attention mechanisms and entirely dispensed with recurrence and convolutions [39]. We propose to adopt the Transformer block-based model instead of PCNN or LSTM since the former outperforms the latter in most NLP tasks owing to the multi-head self-attention mechanism [10]. The second part is a novel attention mechanism, where the sentences that mention the same entity pair are all taken into account by directly utilizing the matching score of each sentence with respect to each relation rather than importing the label relation embedding. Compared with existing neural relation extraction model, our model is expected to better extract the semantic features of each sentence and get more appropriate representation of the bag.

The main contributions of this paper are summarized as follows.

(1) We propose a sentence encoder inspired by Transformer block to effectively conduct feature extraction for each sentence. It not only preserves the consecutiveness of the sentence, but also considers the interactions between different words in the sentence.
(2) We define a novel representation for each bag of sentences by the weighted sentence embedding vectors, where the weight is determined by calculating the probability of each sentence in the bag with respect to each relation in the label set. By doing so, it is expected to achieve more accurate integrated information without additional assistance.
(3) We evaluate the proposed method on a widely used dataset, and the experimental results show that our model is effective for addressing the wrong label problem and significantly outperforms the existing state-of-the-art relation extraction systems. The ablation analyses and the case study further prove the effectiveness and interpretability of our model.

The remainder of the paper is structured as follows. In Section 2, we review the related work on multi-instance learning, DSRE and attention mechanisms adopted in this work. Section 3 presents our hybrid attention-based Transformer block model in detail. In Section 4, the dataset, evaluation metrics, parameter settings and compared algorithms are described. We further analyze our model and compare it with other existing models on the NYT dataset in Section 5. Conclusions and future work are summarized in Section 6.

## 2. Related work

As a fundamental task of text information extraction, relation extraction has been studied extensively, which has mainly applied to various NLP applications, such as automatic text summarization [28], knowledge graph [42], automatic question answering [14] and machine translation [2]. It is designed to extract the relation between two entities from the given text. When two target entities in each sentence have been marked, the purpose is to select a proper relation for these two entities from a predefined relation set. Supervised relation extraction [11,19, 52,55] is always regarded as a multi-class text classification task requiring a large amount of annotated data. In order to address the issue of lacking labeled data in supervised relation extraction, it has been proposed to adopt Freebase to perform DSRE [25], where the basic idea is to automatically label and build datasets. As described in Section 1, the method of data generation sometimes faces the problem of wrong labelling. To alleviate this shortcoming, recent studies have regarded DSRE tasks as a multi-instance learning problem, which was initially proposed when studying the problem of predicting drug activity [18,38]. In multi-instance learning, the main purpose is to distinguish each bag because of the uncertain labels of sentences in it. Here we follow the at-least-one principle, i.e., to regard DSRE as a multi-instance single-label problem.

Nonetheless, the traditional methods [8,13,46] often take a large amount of time to design and verify features that rely on expert knowledge or experience. With the recent huge success of deep learning [4], deep neural network models have been widely used in various applications [9]. Especially, thanks to the ground-breaking work of text classification with CNN models [21], many researchers have investigated the possibility of using neural networks to automatically learn features for relation extraction. An end-to-end CNN based method that automatically captures relevant lexical and sentence level features for supervised relation extraction is proposed in [47]. Subsequently, DSRE has been further improved by Piecewise CNN (PCNN) based on the position of two entities, where for each entity, only the most likely sentence is selected [48]. In other words, among all sentences in the bag that contain the same entity pair, the sentence that best represents the relation is only selected for training and prediction, while the information in other sentences is obviously ignored. To solve the problem, an attention mechanism associated with relation embedding [22] is proposed to utilize the informative sentences in the bag. Based on this, it has been suggested to incorporate the text descriptions of each entity extracted from Freebase and Wikipedia into a similar attention strategy for relation extraction [2]. Besides, the recently proposed Graph Convolution Networks (GCN) [29] have been effectively employed for encoding text information, and dependency trees based GCN model has been used to extract relation [16,50]. Moreover, to reduce the effect of noisy sentences, a nine-layer deep CNN network with residual links [45] is proposed for DSRE, and the generative adversarial network [30] and reinforcement learning [12,31,37] have also been applied for the same sake.

Recently, the attention mechanism has attracted increasing interests [6]. The attention mechanism was first proposed in the field of computer vision and became popular after the Google Mind team used it on the Recurrent Neural Network (RNN) model for image classification [27]. Subsequently, a similar attention mechanism was applied to the machine translation task [2], which was the first time to apply attention mechanism to the field of NLP. With the in-depth study of the attention mechanism, various kinds of attention mechanisms [16,22,23,32,41,53,54] have been proposed for NLP tasks. The Google Machine Translation team received extensive attention in 2017 because of the birth of Transformer model [39], where the multi-head self-attention mechanism plays a major role due to its parallelizability and independencies of distance. Also, many of the subsequent studies are based mainly on the Transformer block, which have been used for various tasks in the field of NLP and achieved the state-of-the-art performance [10,33, 43].

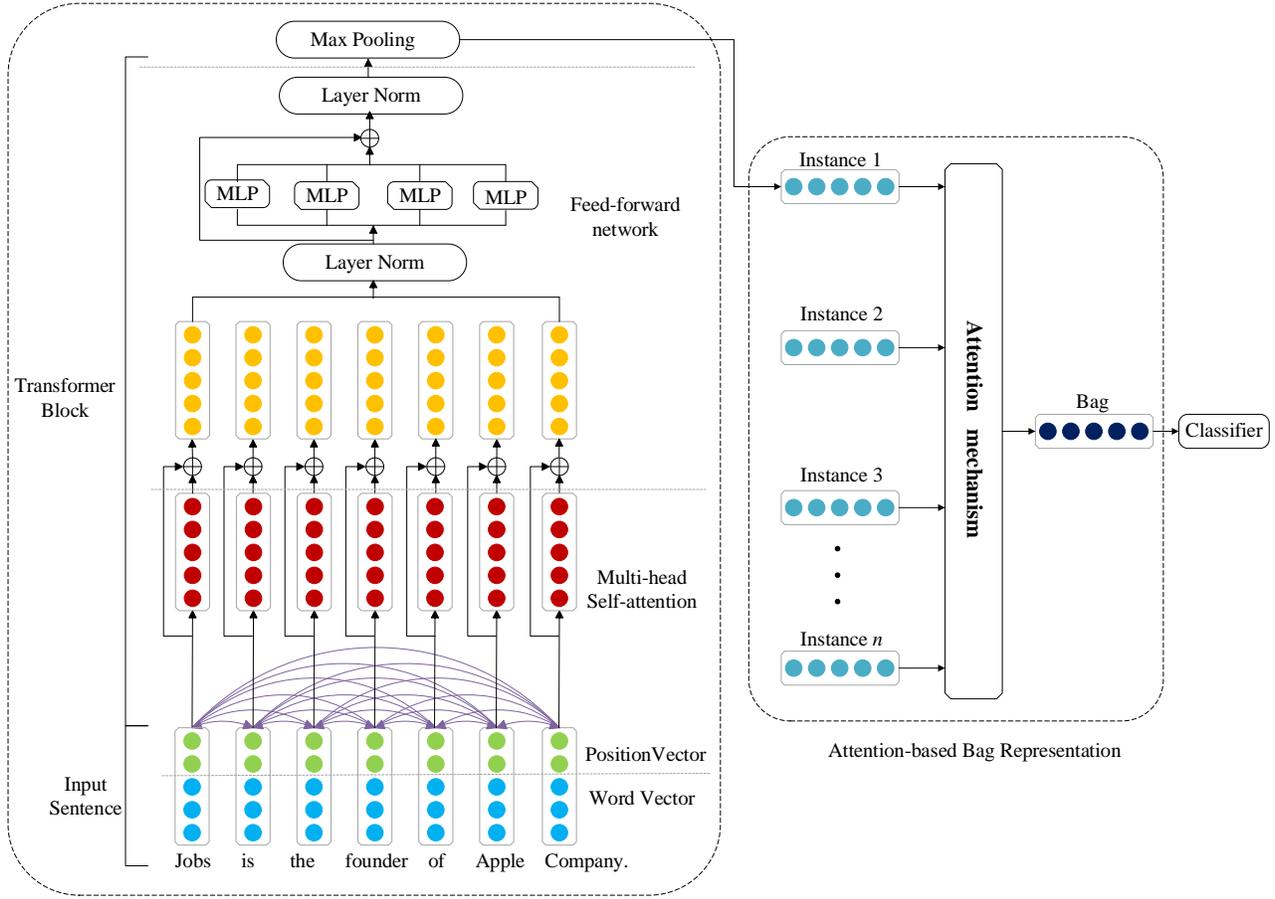

**Fig. 2.** The architecture of our proposed model for DSRE, where the left part illustrates the procedure that handles one instance in a bag and the right part depicts the processing of combining the features extracted from each sentence

## 3. The proposed model

In this section, we present the proposed method for distant supervised relation extraction. For the given instances bag (or sentences bag) that contains the same entity pair $\{S_1, S_2, \cdots, S_n\} \in B$, our model aims to determine the relation from predefined label sets $\{r_1, r_2, \cdots, r_l\} \in R$ by finding out the highest probability as prediction, where $n$ is the number of instances in the bag and $l$ is the number of relations. At first, we perform a vector representation for each sentence in the bag by word embedding and position embedding. Then, the Transformer block-based layer processes the feature extraction over each input sentence vector through a multi-head self-attention mechanism and some other operations. The first two layers form the sentence encoder, which aims to represent the input sentences as vectors and executes feature extraction. After obtaining sentence-level features, the proposed new-type attention-based bag representation is used to get the bag-level features. Eventually, our model learns a group of scores after the above training process, which indicates the confidence that the instances bag belongs to each predefined relation classes. Fig. 2 shows the overall architecture of our model and illustrates the procedure that handles one instance of a bag on the left and the sentence-level attention-based bag representation on the right. In the following, we will describe these two main parts in detail.

### 3.1. Transformer-based Sentence Encoder

For each sentence $S$ containing two target entities in the input layer, Transformer block-based sentence encoder is proposed for automatically learning syntactic features to construct distributed representation of the sentence without complicated preprocessing. The encoder mainly consists of an input layer and a Transformer block-based feature extraction layer. Following the previous work [20,22,48], we also utilize the word embedding and the relative position embedding as the input manner of sentences in relation extraction task. As for the feature extraction layer, there are three main sub-layers in Transformer block-based sentence encoder: a multi-head self-attention layer, a simple position-wise fully connected feed-forward network and a max-pooling layer. The output from each of the first two sublayers is connected by a residual and followed by layer normalization.

*3.1.1 Input Vector Representation*

With the development of deep learning methods for solving the NLP problems, the digital modeling of character symbols is the most critical and primary issue, which aims to convert the text into a distributed representation. In most cases, the inputs of the model are word tokens of sentence, which should be reasonably vectorized to be fed into the network. Therefore, we first transform the text tokens into low-dimensional vectors via a look-up table operation on the word embedding matrix, and then perform fine-tuning in the downstream task. Furthermore, in the relation extraction task, the position information between entities and words is very important. Although the PCNN network can make good use of this point, to some extent, it is at the expense of continuity of the sentence. Hence, the position embedding without segmentation operation is adopted to specify the position of each entity pair in our model.

*Word Embedding*

In general, all parameters of the neural network model are randomly initialized and the training of the network is realized by back-propagation. With the emergence of pre-trained language model [3], experiments show that the model will converge to a better local optimal solution by fine-tuning the pre-trained word vector embedding. Although traditional one-hot word representation method is simple and does not require any training, it is highly memory-intensive and too sparse to cause a "word gap". To capture syntactic and semantic structure of the sentence, we use the skip-gram model [26] to train the word embedding. Given a sentence $S$ consisting of $m$ words $S = \{x_1, x_2, \cdots, x_m\}$, every raw word $x_i$ is represented by a real-valued vector $w_i$ via the pretrained embedding matrix $V \in \mathbb{R}^{d_w \times |V|}$, where $|V|$ denotes the size of corpus vocabulary $V$ and $d_w$ means the dimension of word embedding.

*Position Embedding*

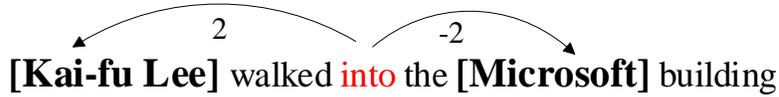

**[Kai-fu Lee]** walked into the **[Microsoft]** building

**Fig. 3.** The relative position of each word and two entities in a sentence

The order of each word appears in the sentence, i.e., the positional features of each word in the sentence is very crucial for the extraction of text information. Especially in the task of relation extraction, the closer the words are to the entities, the more informative the words are. Considering that there are two target entities $e_1$ and $e_2$ in each sentence $S$ and the relative distances to them contribute a lot, here we define the position features (PFs) instead of the positional encodings in Transformer model [39]. With the combination of distance information from each word to two entities, we can get the position vector following the procedure described below. A position embedding matrix $M \in \mathbb{R}^{m \times d_p}$ is first randomly initialized, where $m$ means the length of sentence (numbers of words in the sentence) and $d_p$ is the dimension of position embedding. Then the relative distance of each word in the sentence with respect to the two entities can be calculated as follows:

$$d_i^j = \begin{cases} i - b_j, i < b_j \\ 0, b_j \leq i \leq e_j \\ i - e_j, i > e_j \end{cases} \quad (1)$$

where $i$ is the position of the $i$-th word, $b_j$ means the beginning word of the $j$-th entity and $e_j$ means the ending word of the $j$-th entity. Meanwhile, $j$ equals to 1 or 2, implying the two entities $e_1$ and $e_2$ in the sentence. An example shown in Fig. 3 illustrates how we define the relative distances of each word to two entities. Finally, the position embedding vectors can be obtained by looking up the position embedding matrix according to the calculated relative distances. And two position vectors $p_{1i}$ and $p_{2i}$ can be obtained afterwards.

The most common approach to combining word vectors and position vectors is concatenating, denoted as $x_i = [w_i; p_{1i}; p_{2i}]$, where each word is represented by a real-valued vector $x_i \in \mathbb{R}^d$ and $d = d_w + d_p \times 2$. As shown in Fig. 2, we assume that the dimensions of word embedding and position embedding are 3 and 1, respectively, and consequently the dimension of a word vector is 5. Then, the vector of each word constitutes the representation of the sentence $S \in \mathbb{R}^{d \times m}$, which is processed by a nonlinear function subsequently, e.g., the hyperbolic tangent. Then the final presentation of sentence $S' \in \mathbb{R}^{d \times m}$ is fed into the encoder layer.

$$S' = tanh(S) \quad (2)$$

*3.1.2 Transformer-block-based layer*

Since the introduction of Transformer model in [39], it has successfully achieved promising results in most NLP tasks. The core idea of the attention mechanism used by Transformer is to measure interrelationship of each word with all words in the sentence, thus reflecting the relevance and importance of different words in the sentence to some degree. Therefore, a better representation for each word can be obtained by using these calculated correlations to adjust the weight of each word in the sentence. Since the new representation can imply the word itself and contains the relationship with other words, it is a more global expression than the simple word vector. There are some variations of Transformer block used in relation extraction tasks, including a multi-head self-attention layer, layer normalization, a feed forward layer, another layer normalization and a max-pooling layer.

*Multi-head self-attention mechanism*

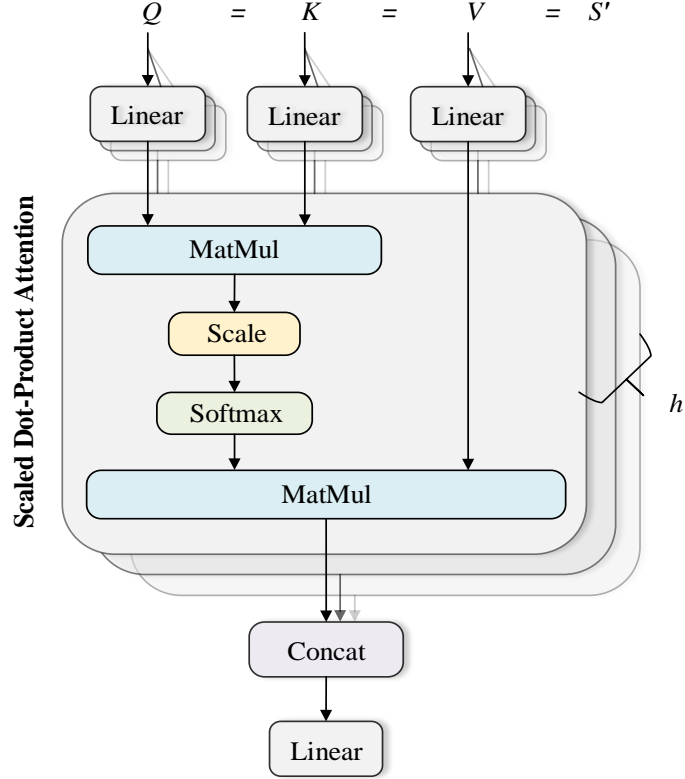

**Fig. 4.** The multi-head self-attention, where the inputs, $Q$(query), $K$(key), and $V$(value) are equal to the representation of sentence $S'$. The sentence-level features can be finally obtained through several linear transformations and scaled dot-product attention.

The multi-head self-attention mechanism plays a crucial role in the Transformer block-based sentence encoder. As illustrated in Fig. 4, there are several linear transformations and scaled dot-product attention in the multi-head self-attention mechanism, which maps a query and a set of key-value pairs to an output. Specifically, we can obtain a weighted sum of the values as output, where the weight assigned to each value is computed by the scaled dot-product attention of the query with the corresponding key. Given the input vectorial variables, queries, keys and values, which are $d_k$-dimensional, $d_k$-dimensional and $d_v$-dimensional, respectively, we first pack them together into matrices $Q$, $K$ and $V$ in sequence to compute the attention function on a set of queries simultaneously. As shown in Eq. (3), the dot products between each query and all keys are computed to measure the correlation between them. Then each dot product is divided by $\sqrt{d_k}$ to avoid too large values. A *softmax* function allows to calculate the weights that are applied to the values to obtain the final representation of attention mechanism. Essentially, the three elements of the model input, query, keys and values are linear transformations of the original input sentence vector if the self-attention mechanism is used.

$$Attention(Q, K, V) = softmax(\frac{QK^T}{\sqrt{d_k}})V \tag{3}$$

Instead of performing a single attention function with the input keys, values and queries, we implement the scaled dot-product attention on a multi-head operation to achieve a more comprehensive presentation. By means of projecting the queries, keys and values $h$ times with different learned linear projections, we can obtain $h$ different linear vectors for each input matrix, where one linear vector corresponds to the input of one head attention mechanism. Eq. (4) defines the calculation method of the $i$-th head. Then the scaled dot-product attention is performed on $h$ parallel heads in order to jointly extract information from different representation subspaces at different positions, as represented in Eq. (5).

$$head_i = Attention(W_i^Q Q, W_i^K K, W_i^V V) \quad (i = 1, \cdots, h) \tag{4}$$
$$MultiHead(Q, K, V) = W^H[head_1, \cdots, head_n] \tag{5}$$
$$A = \{a_1, a_2, \cdots, a_m\} = MultiHead(S', S', S') \tag{6}$$

The parameter matrices $W_i^Q \in \mathbb{R}^{d_{\text{model}} \times d_k}$, $W_i^K \in \mathbb{R}^{d_{\text{model}} \times d_k}$, $W_i^V \in \mathbb{R}^{d_{\text{model}} \times d_v}$ and $W^H \in \mathbb{R}^{d_v \times d_v}$ are learnable for linear transformation. Due to the self-attention as adopted, the input packed queries, keys and values are all equivalent to the input sentence vectors, which can be written as $Q = K = V = S'$. Accordingly, $d_k = d_v = h * d_{\text{model}} = d$ represent the dimension of each vector. The number of packed queries, keys and values is $m$, denoting the number of input words in a sentence. In addition, we set the parameter $h$ to the same value as [39] after many trials and comparisons in relation extraction task. The output of multi-head attention is denoted by Eq. (6), where $a_i \in \mathbb{R}^d$ means the attention-based output vector with respect to the $i$-th word.

*Feed-forward network*

After obtaining the representation of each input word based on the multi-head self-attention mechanism, a feed-forward layer is adopted to better integrate the extracted information, including two linear transformations with a Rectified Linear Units (ReLU) [15] activation in between. There are four Multilayer Perceptrons (MLPs) applied independently to each vector to achieve the transformations, where the number of middle layer neurons is three times larger than the embedded dimension. Regarding the number of MLPs, we have conducted a couple of trials and found that the setting of 4 is moderate. Besides, the motivation of using ReLU is due to its advantage of sparse representation, which can reduce the interdependence between parameters and alleviate the overfitting problems. To be specific, the variable $a_i$, corresponding to the $i$-th position in $A$, is processed by the feed-forward network and remains the original size:

$$f_i = FFN(a_i) = Re\,L\,U(a_i W_1 + b_1) W_2 + b_2 \tag{7}$$

$$Re\,L\,U(x) = \begin{cases} x, x \geq 0 \\ 0, x < 0 \end{cases} \tag{8}$$

where $W_1 \in \mathbb{R}^{d \times 4d}$, $W_2 \in \mathbb{R}^{4d \times d}$ are the transformation matrices, $b_1, b_2$ are bias terms, and $f_i \in \mathbb{R}^d$ imply the sentence vector representation after the feed-forward network layer.

In order to keep both the weight scale invariance and data scale invariance to better train the network model, the normalized layer is employed to merely act on the embedded dimension on a single training case through computing the mean and variance of the data. The output of the above two sub-layer is finally connected by a residual connection [17] and followed by layer normalization [24] to further reduce the training time, given as:

$$O = LayerNorm(X + FNN(X)) \tag{9}$$

where $X = LayerNorm(S' + A)$ and $O \in \mathbb{R}^{d \times m}$.

*Max-pooling*

Since the generated sentence-level feature vectors should be the same size for each input sentences, a rational methodology is used to merge such extracted features due to the various sentence lengths. In addition, max-pooling is more capable of identifying the most important and relevant features from the sequence than the average-pooling. The final sentence presentation is described in Eq. (10), where we can see that the dimension of sentence vector is no longer related to the sentence length:

$$P = max\{O\} = max\{o_1, o_2, \cdots, o_m\} \tag{10}$$

As a result, the sentence-level feature of $S' \in \mathbb{R}^{d \times m}$ is represented as $P \in \mathbb{R}^d$.

*3.2. Attention-based Bag Representation*

As mentioned earlier, DSRE always suffers from the issue of wrong labelling. To reduce the interference of noise sentences while utilizing all the valid sentences, a novel sentence-level attention mechanism base on the work in [22] is proposed. Given the distributed representation of each sentence, the attention mechanism is modified without any auxiliary operation.

The vector representation of sentence $S_i$ in a certain instances bag $B$ can be expressed as $P_i \in \mathbb{R}^d$, which is obtained from Transformer block-based sentence encoder in Section 3.1. In order to better represent each bag, we merge the information of each sentence as shown in Fig. 5. The max-pooling layer is followed by a fully connected network to obtain the confidence of sentence $S_i$ associated with relation $r_k$, denoted as $u_{ik}$. Specifically, the $k$-th component of $U_{ij}$ is formulated as:

$$U_{ij} = W_3 P_i + b_3 \tag{11}$$

where $W_3 \in \mathbb{R}^{l \times d}$ is also the transformation matrix and $U_i \in \mathbb{R}^l$ stands for the final relation vector of input sentence, where $l$ means the number of pre-defined relation labels.

To alleviate the influence of wrong labelled sentences in the bag, the representation of instances bag is obtained by directly calculating the contribution of each relation of each sentence to the instances bag without introducing any other embedding. The confidence of instances bag $B$ related to $r_k$ (denoted as $b_k$) is calculated according to $\alpha_{ik}$, indicating the contribution ratio of sentence $S_i$ to instances bag $B$:

$$b_k = \sum_{i=1}^{n} \alpha_{ik} P_i \quad (k = 1, \cdots, l) \tag{12}$$

$$\alpha_{ik} = \frac{exp(u_{ik})}{\sum_{i=1}^{n} exp(u_{ik})} \tag{13}$$

where $n$ is the number of sentences in the bag.

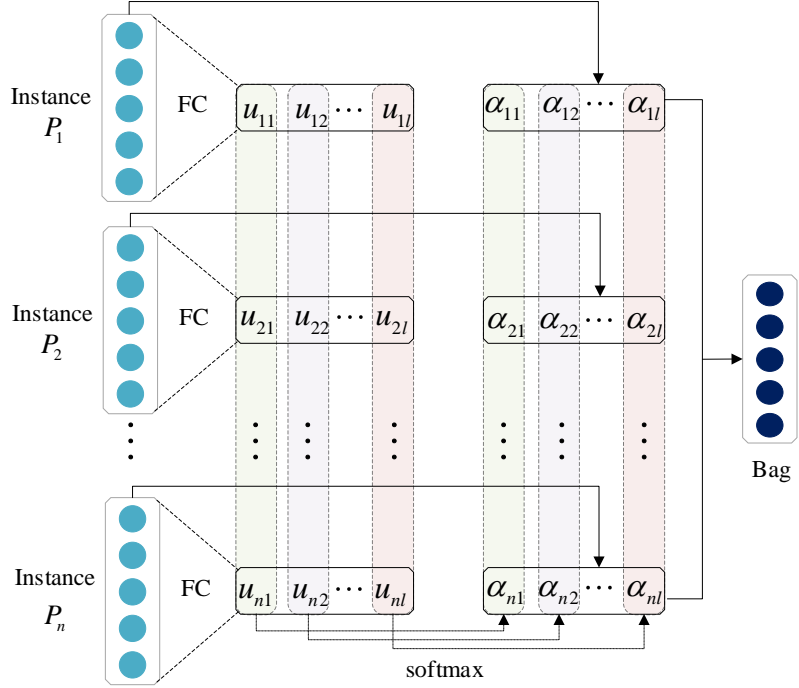

**Fig. 5.** Attention-based bag representation. It is assumed that the number of relation labels is $l$, and there are $n$ instances in the bag. We merge the information to better represent the bag by calculating the contribution of each sentence to the instances bag.

Combined with Fig. 5, the scores of each instance $u_{ik}$ are first received through the fully connected network, and then the *softmax* function is operated on the scores according to the same label dimension. Next, the calculated weight values $\alpha_{ik}$ are obtained, some of which are expected to assign smaller values to reduce the contribution of the noisy sentences to the bag. To sum up, the matching degree of each bag to each label is estimated according to the matching degree of each sentence in the bag to each label. This way, we alleviate the issue of introducing label embedding that may cause errors.

*3.3. Classification and Lost function*

The relation vector of instances bag $B$ can be obtained through our model. To obtain the conditional probability that $B$ is classified into relation $r_k$, we apply the classifier over all relation types:

$$p(r_k|B) = \frac{exp(b_k)}{\sum_{k'}^{l} exp(b_{k'})} \qquad (14)$$

For each bag, the predicted relation corresponds to the class with the highest probability. Suppose that the training set including $T$ instance bags $\{B_1, B_2, \cdots, B_T\}$ and $(B_i, r_i)$ is the *i*-th sample, we define the objective function using cross-entropy at the bag level as follows:

$$J(\theta) = -\sum_{i=1}^{T} \log p(r_i|B_i; \theta) \qquad (15)$$

where $\theta$ contains the whole learnable parameters. The stochastic gradient descent (SGD) optimizer is exploited to minimize the loss function and the dropout method [36] is applied after word embedding and sentence encoder to prevent co-adaptation of hidden units by randomly omitting feature detectors through an element-wise multiplication.

## 4. Experiments

*4.1. Dataset and Evaluation Metrics*

The proposed model architecture is evaluated by the NYT dataset (http://iesl.cs.umass.edu/riedel/ecml/), which is widely used in verifying the performance of the recently proposed DSRE algorithms. This dataset was first released by [34] and generated automatically by aligning Freebase relations with the NYT corpus, where sentences from 2005-2006 are used as training instances, and sentences from 2007 are used as testing instances. Besides, entity mentions are annotated using the Stanford named entity tagger [35] and further matched to the names of Freebase entities. There are 53 possible relationship labels including a special relation NA, indicating that there is no relation between the entity pair in the instance. Statistics of the datasets is summarized in Table 1. Relational facts refer to the number of sentences whose labels are not NA.

Since the dataset is based on the assumption of distant supervision, it signifies that the labels of the sentences containing the same entity pairs are the same. To be consistent with others, we package sentences with the same entity pairs into a bag to obtain the bag-level samples. Assuming that the data distribution of the training set is similar to the validation set, our model is evaluated on the held-out test set of NYT dataset. The held-out evaluation provides an effective way to assess the model by comparing the predicted relations against those in Freebase. The Precision/recall (PR) curves and Precision@N (P@N) values are adopted as the metric in each set. Since most of the test instances bags contain only one sentence, which is related to the bag representation, we divide all the test sets into three levels based on the number of sentences in each bag.

**One**: For each entity pair, we randomly select one sentence to form the instances bag.
**Two**: For each entity pair, we randomly select two sentences to form the instances bag.
**All**: For each entity pair, we select all sentences to form the instances bag.

**Table 1**
Details of datasets in the paper.

|  | Sentence | Entity pairs | Relational facts | Labels |
| --- | --- | --- | --- | --- |
| Train | 570088 | 281270 | 18252 | 53 |
| Test | 172448 | 96678 | 1950 | 53 |

### 4.2. Parameter Settings

**Table 2**
Parameters setting of the model.

| Parameter | Value |
| --- | --- |
| Word embedding dimension $d_w$ | 50 |
| Position embedding dimension $d_p$ | 5 |
| Heads $h$ | 8 |
| Batch size | 100 |
| Learning rate | 0.05 |
| Dropout probability | 0.5 |

In this work, we use the word2vector (https://code.google.com/p/word2vec/) to train the word embedding that appeared more than 100 times in the NYT corpus, and the vocabulary size of constructed word dictionary is 114044. Additionally, the position embedding is randomly initialized with a uniform distribution between $[-1,1]$. The 50-dimensional trained word embedding and two 5-dimensional position embedding are concatenated to constitute the whole input word vector. All hyperparameters used in our model are listed in Table 2, most of which are set empirically. The mini-batch SGD with the initial learning rate of 0.05 is employed for optimization. During the training process, the learning rate is iterated every 20 epochs with a decay rate of 0.1.

### 4.3. Compared algorithms

The following existing algorithms are selected to verify the effectiveness of our proposed methods through held-out evaluations.
**Mintz** [25]: A traditional multi-class logistic regression model.
**MultiR** [18]: A probabilistic, graphical model of multi-instance learning.
**MIML** [38]: A multi-instance multi-label model.
**PCNN** [48]: A piecewise-CNN based model.
**PCNN+LABLE** [22]: A relation embedding based selective attention mechanism with PCNN model.
**BGWA** [20]: A Bi-GRU based model with word and sentence level attention.

For the proposed model, another two kinds of methods are designed to verify the respective function of Transformer block-based sentence encoder and sentence-level attention based bag representation.
**PCNN+SA**: Sentence-level attention based PCNN model.
**Trans+LABLE**: A relation embedding based selective attention mechanism with Transformer block sentence encoder.

# 5. Results and discussions

## 5.1. Performance Comparison

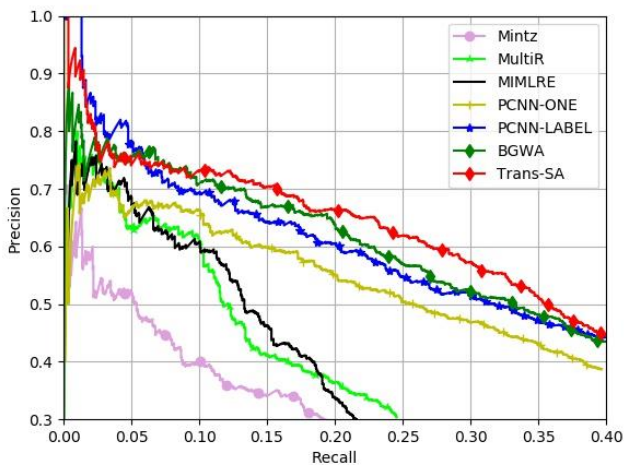

**Fig. 6.** Precision-recall curves of the proposed model and compared algorithms.

**Table 3**
Precision of various models for different recalls.

| Model | Recall | | | |
|---|---|---|---|---|
| Precision(%) | 0.1 | 0.2 | 0.3 | 0.4 |
| Mintz | 39.9 | 28.6 | 16.8 | - |
| MultiR | 60.9 | 36.4 | - | - |
| MIML | 60.7 | 33.8 | - | - |
| PCNN-ONE | 66.7 | 55.6 | 47.8 | 39.5 |
| PCNN-LABEL | 69.6 | 60.1 | 51.6 | 45.2 |
| BGWA | 71.2 | 63.9 | 51.8 | 45.3 |
| Trans-SA | **74.1** | **67.2** | **57.9** | **45.4** |

**Table 4**
P@N for relation extraction in the entity pairs with different number of test sentences.

| P@N (%) | One | | | | Two | | | | All | | | |
|---|---|---|---|---|---|---|---|---|---|---|---|---|
| | 100 | 200 | 300 | Mean | 100 | 200 | 300 | Mean | 100 | 200 | 300 | Mean |
| Mintz | 35.0 | 37.5 | 37.3 | 36.6 | 51.0 | 42.0 | 43.3 | 45.4 | 54.0 | 50.5 | 45.3 | 49.9 |
| MultiR | 64.0 | 61.5 | 53.7 | 59.7 | 62.0 | 61.5 | 58.7 | 61.1 | 75.0 | 65.0 | 62.0 | 67.3 |
| MIML | 62.0 | 59.0 | 54.7 | 58.6 | 69.0 | 59.5 | 59.0 | 62.5 | 70.0 | 64.5 | 60.3 | 64.9 |
| PCNN-ONE | 73.3 | 64.8 | 56.8 | 65.0 | 70.3 | 67.2 | 63.1 | 66.9 | 72.3 | 69.7 | 64.1 | 68.7 |
| PCNN-LABEL | 73.3 | 69.2 | 60.8 | 67.8 | 77.2 | 71.6 | 66.1 | 71.6 | 76.2 | 73.1 | 67.4 | 72.2 |
| BGWA | 78.0 | 71.0 | 63.3 | 70.7 | 81.0 | 73.0 | 64.0 | 72.6 | 82.0 | 75.0 | 72.0 | 76.3 |
| Trans-SA | **84.0** | **76.5** | **70.0** | **76.1** | **86.0** | **78.5** | **73.0** | **79.1** | **84.0** | **81.0** | **79.6** | **81.5** |

In this section, we compare our model with peer methods. The precision-recall curves are shown in Fig. 6, and Table 3 displays the precision of these models at different recalls (0.1/0.2/0.3/0.4) to give a more explicit comparison. Overall, we can make the following observations.
1）Compared with the non-neural-network models, neural network models improve the performance significantly as the performance of the former deteriorates quickly when the recall is greater than 0.1, while the latter can maintain a reasonable performance in the meantime. The is due to the fact that most of the features used in non-neural-network models come from NLP tools, such that the accumulation of errors can lead to degraded model performance.

2）Furthermore, we can see that our model achieves an improvement in precision, especially at middle recall levels, where the precision is 60% with recall of 27%. By contrast, the recalls are decreased to approximately 22% and 20%, respectively, when PCNN-LABEL and BGWA achieve the same precision. This improvement is attributed to not only the Transformer block-based encoder, but also the sentence-level attention-based bag representation. It is worth noting that our model does not require any artificial features and external knowledge, where the input is the complete sentence without any complicated data preprocessing, such as mask or piecewise operation. The Trans-SA model outperforms all the compared systems and achieves higher precision over most recall range even in such cases.

Following the previous work [20, 22], we also evaluate our method with different numbers of sentences and the results in three test settings are summarized in Table 4. Among those entity pairs that contain at least two sentences, we randomly select one, two or all sentences to form the test set. The metric P@N indicates the precision of the relation extraction results with the top N highest probabilities in the held-out test set. And here we report the P@100, P@200, P@300 and the mean of them for each model. As can be seen from Table 4, our proposed methods achieved higher P@N values than previous works, and the results get better as the number of sentences in each bag increases. It implies that taking more complete and useful information into account will be more reliable for the relation extraction of each bag.

*5.2. Ablation Analyses*

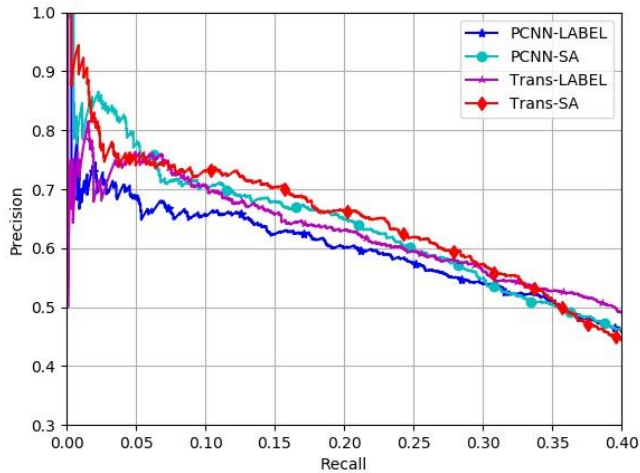

**Fig. 7.** Precision-recall curves of models with different settings

In order to separately prove the validity of each part of our model, we empirically perform ablation analyses through held-out evaluation by a pairwise combination on each part of Trans-ALL and PCNN-LABEL. Fig. 7 depicts the precision-recall curves of models that combine different sentence encoders and bag representations, where the higher performance of Trans-SA and Trans-LABEL over PCNN- SA and PCNN-LABEL indicates that our sentence encoder is more beneficial to the feature extraction of each sentence. Meanwhile, by comparing the two groups of experiments using PCNN-LABEL with PCNN-SA and Trans-LABEL with Trans-SA, we can verify effectiveness of the novel attention mechanism. As a result, it outperforms other models in the same test settings regardless of whether the encoder is Transformer block or PCNN, indicating that our improved attention mechanism helps more in DSRE.

*5.3. Case Study*

In this subsection, we will take sentences in the dataset as an example to illustrate the performance of each part of our model to make it interpretable. Fig. 8 shows the visualization of multi-head self-attention mechanism inside the sentence encoder to demonstrate that our model is useful. It measures the correlation between different words in a sentence and two entities, by which we can obtain the richer word representations. The color density indicates how much an entity focuses on each word in a sentence. For the two entities in the given sentence, the related words are different and displayed in shades of color, where heavier colors indicate more relevant words. In addition to the context words of the second entity "*Brussels*", the other most prominent words are "*European*" and "*in*", from which we can recognize that the relation may be relevant to the geographical location and the entity pair is commonly concentrated before and after "*in*". The reasoning process is the same to the second entity. Actually, these words play the most important role in semantically predicting "*/location/location/contains*", which is the relation class of this entity pair.

Table 5 mainly shows that how we specifically fuse the sentences in each bag and figure out the relation label via the sentence-level attention mechanism. In the fourth column of the table, the number of each group respectively indicate "the rank of the probability" that the sentence in the bag is classified into the ground truth label and "the contribution of this sentence to the bag" (i.e., $\alpha_{ik}$ in Eq. (12)). The array elements in the second column also imply "the rank of the probability" and "the probability" that the bag is classified into the ground truth label, where for the former, '1' meaning correct classification, and for the latter, the variable is $p(r_k|B)$) in Eq. (14). With respect to the relational triple <*Minnesota, /location/location/contains, Mankato* >, there are two sentences in the bag, where the first sentence is assigned with higher contribution since it indicates that "*Mankato*" is a state in "*Minnesota*". Moreover, the results are reasonably distributed and the bag is also eventually classified correctly. Therefore, the sentence-level attention mechanism is capable of learning more valid information for the bag presentation.

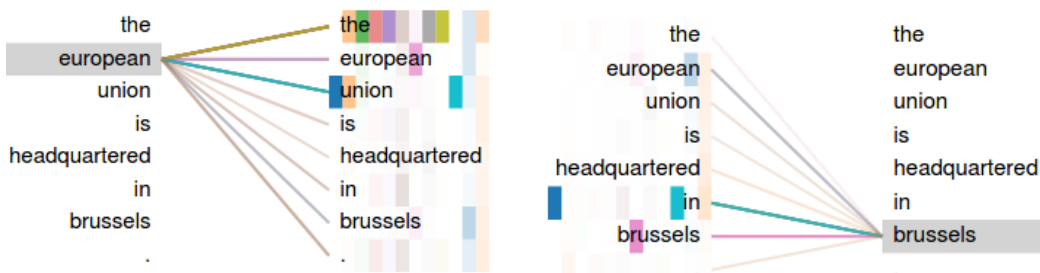

**Fig. 8.** Visualization of multi-head self-attention mechanism. Heavier the colors indicates more relevant words

**Table 5**

Some examples of sentence-level attention for bag presentation.

| Relation and Entity pair | (rank, probability) | Bag | (rank, $\alpha_{ik}$) |
|---|---|---|---|
| /location/location/contains （Minnesota, Mankato） | (1, 9.9685e-01) | ……her family survives her in **Mankato**, **Minnesota**…… | (1, 0.9994) |
| | | ……winona trailed by 16 points with 8 minutes 8 seconds left before rallying to beat **Minnesota** state **Mankato** in regulation , 74-71…… | (53, 0.0006) |
| /people/person/nationality (Shashi_tharoor, India) | (1, 7.1322e-01) | ……he said friday that he had accepted 17 of resignations , one of which was from **Shashi_tharoor** of **India**…… | (4, 0.4022) |
| | | …... **Shashi_tharoor** 's goal was to make me feel stupid for preferring baseball to cricket , he would have been better served had he not included the following : " cricket is better suited to a country like **India**…… | (27, 0.3632) |
| | | ……op-ed contributor **Shashi_tharoor**, a departing under secretary general of the united nations , is the author of " the great **Indian** novel "…… | (48, 0.2346) |

## 6. Conclusion and future work

In this paper, we have proposed the Trans-SA, a novel neural network model to automatically learn features from each sentence and DSRE. The traditional PCNN divides the sentence into three segments, which may lead to a lack of semantic globality, we adopt the Transformer block for encoding syntactic information of sentences. In addition, a novel improved attention mechanism is presented to deal with the wrong labelling problem. Through extensive experiments on the widely used dataset, we demonstrate the competitive performance of our model in comparison with the state-of-the-art peer methods. In addition, the ablation analyses have indicated the effectiveness of each component in the proposed model. Essentially, the proposed model can be used as a tool for DSRE with the entities marked.

In the future, it would be interesting to apply the proposed framework to implement multiple tasks, such as the joint extraction of named entity recognition (NER) and relation extraction (RE). At the same time, we can also try to integrate some external linguistic knowledge into our model, which can be believed to enhance performance and make deep neural network more preferable for natural language processing tasks. Further, we consider applying our methods to other domains such as biomedical or scientific articles in order to further benefit this task.